\documentclass[a4paper, 10pt, conference]{ieeeconf}  
\IEEEoverridecommandlockouts

\usepackage{amsmath,amsfonts}
\usepackage{graphics} 
\usepackage{adjustbox}
\usepackage{booktabs} 
\usepackage{hyperref}
\usepackage{subcaption}

\DeclareMathOperator*{\argsort}{arg\,sort}
\DeclareMathOperator*{\argmin}{arg\,min}

\newcommand{\CP}[1]{\ignorespaces}
\usepackage{flushend}

\newcommand{\PC}[1]{\ignorespaces}
\newcommand{\ie}{\textit{i.e.\, }}
\newcommand{\eg}{\textit{e.g.}}

\title{\LARGE \bf TReR: A Lightweight Transformer Re-Ranking Approach \\ for 3D LiDAR Place Recognition}

\author{Tiago Barros,  Lu\'{i}s Garrote, Martin Aleksandrov, Cristiano Premebida, Urbano J. Nunes 
\thanks{T. Barros, L. Garrote, C. Premebida, and U.J. Nunes are with the University of Coimbra, Institute of Systems and Robotics, Department of Electrical and Computer Engineering, Portugal. E-mails:\{tiagobarros,~garrote,~cpremebida,~urbano\}@isr.uc.pt. Martin Aleksandrov is with Dahlem Center for Machine Learning and Robotics, Freie Universit\"{a}t Berlin, Berlin. E-mail: martin.aleksandrov@fu-berlin.de.}%
}
\begin{document}

\maketitle
\thispagestyle{empty}
\pagestyle{empty}
\begin{abstract}
Autonomous driving systems often require reliable loop closure detection to guarantee reduced localization drift. Recently, 3D LiDAR-based localization methods have used retrieval-based place recognition to find revisited places efficiently. However, when deployed in challenging real-world scenarios, the place recognition models become more complex, which comes at the cost of high computational demand. This work tackles this problem from an information-retrieval perspective, adopting a first-retrieve-then-re-ranking paradigm, where an initial loop candidate ranking, generated from a 3D place recognition model, is re-ordered by a proposed lightweight transformer-based re-ranking approach (TReR). The proposed approach relies on global descriptors only, being agnostic to the place recognition model. The experimental evaluation, conducted on the KITTI Odometry dataset, where we compared TReR with s.o.t.a. re-ranking approaches such as $\alpha QE$ and SGV, indicate the robustness and efficiency when compared to $\alpha QE$ while offering  a good  trade-off between robustness and efficiency when compared to SGV.      
\end{abstract}
\section{INTRODUCTION}\label{sec:introduction}

Place recognition has become a popular approach to perform global localization and loop closure detection~\cite{cattaneo2022lcdnet}. With the development of efficient deep learning (DL) techniques able to model geometrical information, 3D LiDAR information has gained increasing relevance in place recognition due to being more robust compared to visual data in appearance-changing environments: a feature of key importance in the long-term operation of autonomous vehicles~\cite{barros2021place}. In 3D LiDAR-based place recognition, DL networks are used to map the geometrical information (\ie point clouds) into an adequate descriptor space, where descriptors from nearby physical places are close. In contrast, descriptors from farther physical places are apart. Nevertheless, as the DL-based models are exposed to more realistic scenarios, the more complex they become, which usually means more parameters to train and, consequently, more computational power is required for training. In other fields, such as information retrieval (IR), a common approach to this problem is the first-retrieve-then-re-ranking (RTrR) paradigm~\cite{tan2021instance,zhang2023etr}. 


\begin{figure}[h]
    \centering
    \includegraphics[width=\linewidth]{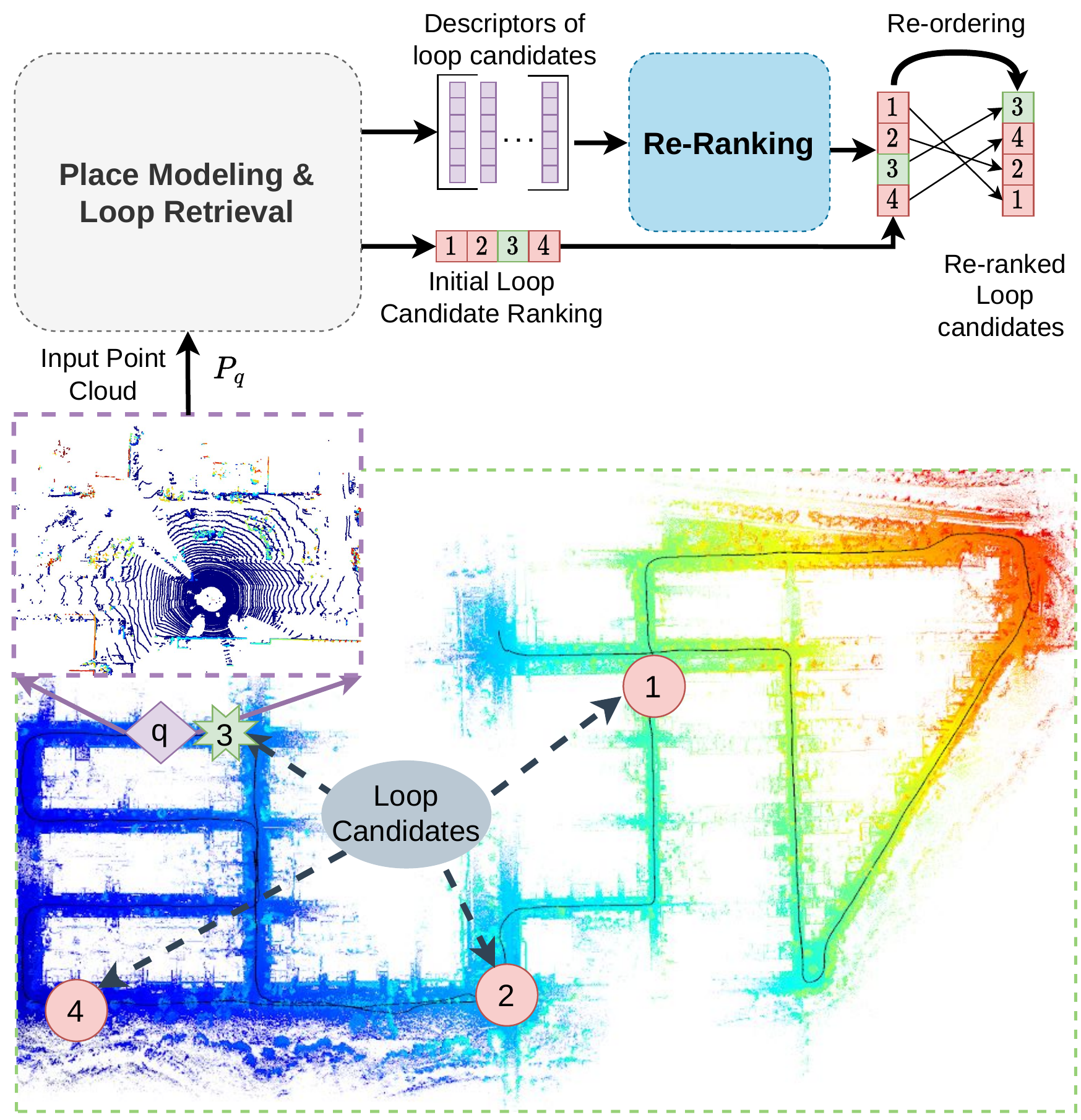}
    \caption{A two stage framework with 3D LiDAR-based place recognition and re-ranking. For a given query $q$ location, a pre-trained place recognition model first maps the point cloud $P_q$ to a descriptor. The descriptor is used to query a database to return the $k$ nearest neighbors, which represent an initial loop candidate ranking. The ranking is ordered from the most to the least similar.  This process is illustrated with the following example: for the query point cloud $P_q$ captured from the location $q$, four loop candidates are returned, which are identified in the map by a number and a shape. The number represents the initial rank, and the shape represents true loops (green star) or false loops (red circle).}
    \label{fig:fronte}
\end{figure}

In an RTrR architecture, two sequential stages take place: an initial coarse retrieval stage, where pre-trained models are used, and a re-ranking stage, where smaller re-ranking models are used, trained on a target domain dataset~\cite{chen2022deep}. In the first stage, the goal is to learn global descriptors from the input point clouds. In the second stage, the goal is to learn interactions~\cite{INR-100}, associating a higher rank to more relevant candidates and a lower rank to less relevant candidates, such as false positives, as a way to refine the initial estimate. In traditional loop closure detection, false positives are identified using geometrical verification approaches such as RANSAC~\cite{fischler1981random} and Iterative Closest Point~\cite{zhang1994iterative}, which are computationally demanding and, for this reason, often not being well-suited for large-scale applications. 

In this work, we study the viability of the first-retrieve-then-re-ranking paradigm in place recognition, as shown in Figure~\ref{fig:fronte}.
Inspired by the success of re-ranking-based applications in image retrieval~\cite{tan2021instance,zhang2023etr}, our approach, called TReR, is a lightweight, transformer-based model trained in a supervised regime. 
Unlike common transformer-based approaches for vision applications (\eg ~\cite{tan2021instance,zhang2023etr}), which rely on local features and return similarity scores for pairs of images, TReR relies on global descriptors only, making it lightweight and agnostic to place recognition models. Furthermore, TReR computes similarity scores for $k$ loop candidates simultaneously instead of pairs.

We assess the adequacy of the approach within a 3D LiDAR place recognition framework for urban contexts. TReR is evaluated on five different place recognition models~\cite{uy2018pointnetvlad,vidanapathirana2022logg3d,
barros2023orchnet}, which serve as baselines. Additionally, we also compare TReR to another global descriptor-based re-ranking approach ($\alpha QE$ \cite{radenovic2018fine}) and a geometrical verification (GV) method (SGV\cite{10065560}). The results are obtained on the KITTI dataset and indicate that TReR achieves a greater performance than the baselines (with no re-ranking) and $\alpha QE$ while offering an acceptable trade-off between performance and efficiency in contrast to SGV\cite{10065560}.
In brief, the main contributions of this work are the following: 
\begin{itemize}
    \item An lightweight Transformer-based re-ranking approach (TReR) for 3D LiDAR place recognition that relies only on global descriptors and predicts the relative similarity scores of $k$ candidates in one forward pass; 
    \item  TReR improves retrieval performances compared to other global-based methods and offers a good compromise between performance and efficiency.
\end{itemize}

 \begin{figure*}[t]
    \centering
    \includegraphics[width=\textwidth]{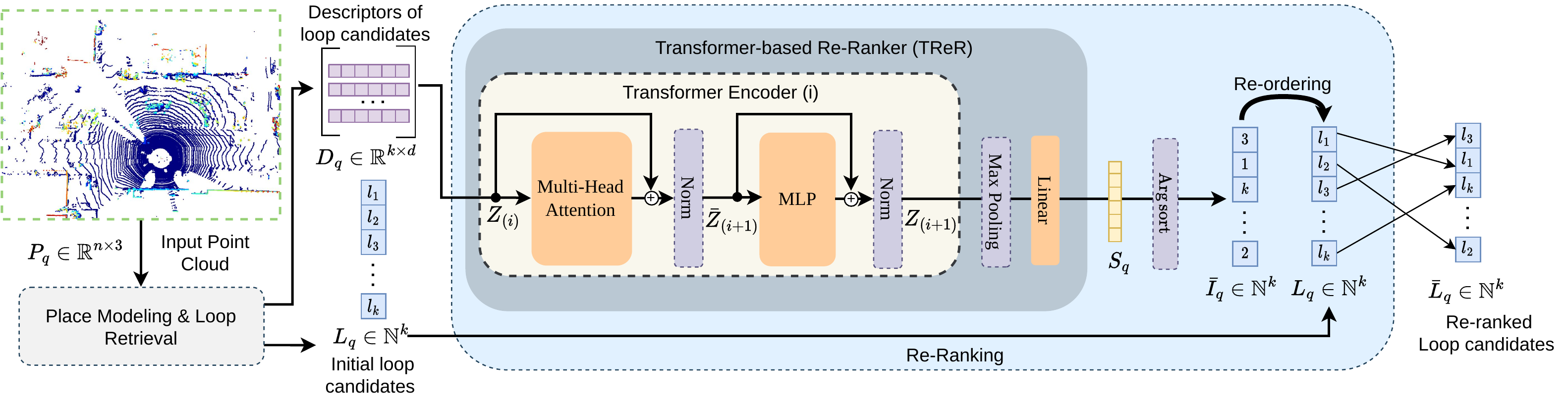}
    \caption{The place recognition framework with two sequential tasks: (1) place modeling (Loop retrieval) and (2) re-ranking. The first model (Loop retrieval) receives a query point cloud $P_q$ and then returns a loop ranking $L_q$ and the corresponding descriptors $D_q$. The proposed re-ranking approach TReR receives the descriptors as input and returns a new ranking $\bar{L}_q$ as output.}
    \label{fig:framework}
\end{figure*}

\section{RELATED WORK}
Place recognition has been an active field of research over the last decade, used in increasingly complex scenarios such as long-term operation in ever-changing environments~\cite{barros2021place}. The current endeavor to deal with such scenarios has been in the direction of proposing more complex models, requiring higher computational resources, which are not easily available. In other fields, such as IR, a common approach is to use an RTrT architecture, where, in the first stage, pre-trained models are used to generate coarse ranking estimates, and in the second stage, these initial estimates are refined using re-ranking models~\cite{chen2022deep}. 

We next outline and discuss works on 3D LiDAR-based place recognition, which relate to the first stage, and works on re-ranking, which relate to the second stage.

\subsection{3D-LiDAR-based Place Recognition}\label{subsec:threedlidar}

In 3D LiDAR-based place recognition, works such as PointNetVLAD~\cite{uy2018pointnetvlad} and LPD-Net~\cite{liu2019lpd} pioneered the idea of modeling physical places from point clouds directly by using end-to-end frameworks. Such frameworks, similar to those for vision-based place recognition, extract local features from point clouds and then aggregate the local features into a global descriptor. Over the years, the research interests have focused on different aspects of place recognition frameworks. 

Some works focus on improving the feature extraction from point clouds \eg, addressing the rotation invariance problem in neural networks~\cite{9712221}, proposing more suitable feature extraction operators for point clouds, such as sparse 3D convolutions~\cite{vidanapathirana2022logg3d} and hierarchical graphs~\cite{shu2023hierarchical}. 

Other works have focused on aggregating local features into global descriptors. Most aggregators were initially proposed for images, such as the popular NetVLAD~\cite{arandjelovic2016netvlad}, which is a trainable generalization of VLAD~\cite{jegou2010aggregating}. Other approaches, for instance, resort the attention to re-adjusting the weights of local features \cite{barros2022attdlnet}, or compute various pooling operations on the local features, such as sum pooling (SPoC)~\cite{babenko2015aggregating} or generalized-mean pooling (GeM)~\cite{radenovic2018fine}. In addition, the combination of the aforementioned aggregators into one global aggregator to leverage the strength of each aggregator has been explored in~\cite{barros2023orchnet}. 

In this work, we use PointNetVLAD~\cite{uy2018pointnetvlad},  LoGG3D-Net~\cite{vidanapathirana2022logg3d}, and ORCHNet~\cite{barros2023orchnet} for generating global descriptors and, additionally, we have adapted SPoC~\cite{babenko2015aggregating} and GeM~\cite{radenovic2018fine} for working with point clouds.

\subsection{Re-Ranking}\label{subsec:rerankinglit}

A re-ranking approach takes an initial ranking estimate and re-orders it based on a pre-defined criterion. In information retrieval, such a criterion is to rank higher content that is more relevant for a given query. In retrieval-based place recognition, the aim is to promote loop candidates that are true loops while ranking lower false positives. 

These re-ranking methods can be categorized into two families: those that use only global descriptors and those that use local and global features combined. As for the global descriptors only, a widely used approach is Query Expansion (QE). In particular, $\alpha QE$ has been used both in image retrieval~\cite{radenovic2018fine} and retrieval-based place recognition~\cite{10065560}. On the other hand, the category of methods that use both local and global information is more common in applications with geometrical data. For instance, RANSAC~\cite{fischler1981random} and Iterative Closest Point~\cite{zhang1994iterative} are widely adopted in localization-related tasks, where they are employed to assess how well two distinct point clouds match. These methods, known as geometrical verification (GV), work well when seeking false positives but are usually computationally demanding when working with large point clouds. More recently, a more efficient approach was proposed in \cite{10065560} called spectral geometrical verification (SGV), which relies on correspondence compatibility graphs~\cite{leordeanu2005spectral} to assess the spatial consistency of two point clouds.        


\section{PROPOSED APPROACH}\label{sec:approach}

In this section, we describe the proposed \textbf{T}ransformer-based \textbf{Re}-\textbf{R}anking framework (TReR) outlined in Figure~\ref{fig:framework}. Firstly, we formulate the place recognition and re-ranking tasks in generic terms. Secondly, we present the proposed approach in more detail.

\subsection{Problem Formulation}\label{subsec:problem}

The proposed framework is depicted in Figure~\ref{fig:framework}. It can be split into two sub-tasks: 1) loop retrieval, using a place recognition approach for generating $k$ candidates; 2) loop re-ranking, which re-orders the initial loop ranking, giving greater ranks to the true loops and lower ranks to the false loops.

\subsubsection{Loop Retrieval} \label{subsubsec:retrieval}

A point cloud-based loop retrieval task can be formulated as follows: given a query 3D point cloud $P_q \in \mathbb{R}^{n\times 3}$ with $n$ points, and a database with descriptors of already visited places, the task goal is to map $P_q$ to a descriptor vector $d_q \in \mathbb{R}^{d}$ of $d$ dimensions, and use $d_q$ to query the database to retrieve the $k$ nearest neighbors based on a similarity score computed using the Euclidean distance. The retriever returns a set of database indices  $L_q = \{l_i\}_{i=1}^{k} \in \mathbb{N}^k$, where $l_i$ is the index (\ie position in the database) of the $i$-th retrieved loop candidate. The candidates are ordered from the most similar to the least similar.

In this work, a loop candidate $l_i \in L_q$ is a true loop, when the loop candidate is within a given $\zeta$ Euclidean distance from the query: \ie both the query and the respective candidate originated from the same place. In generic terms, this can be formulated by the following function: 

\begin{equation}
    \Gamma:\mathbb{R}^{n\times 3} \rightarrow \mathbb{N}^{k}\mbox{,}
\end{equation}

\noindent where $\Gamma$ can be learned. Since this work focuses on re-ranking, we refer the reader to \cite{8578568} for more details on the training procedure for this learning task.

\begin{figure}[!t]
    \centering
    \includegraphics[width=\linewidth]{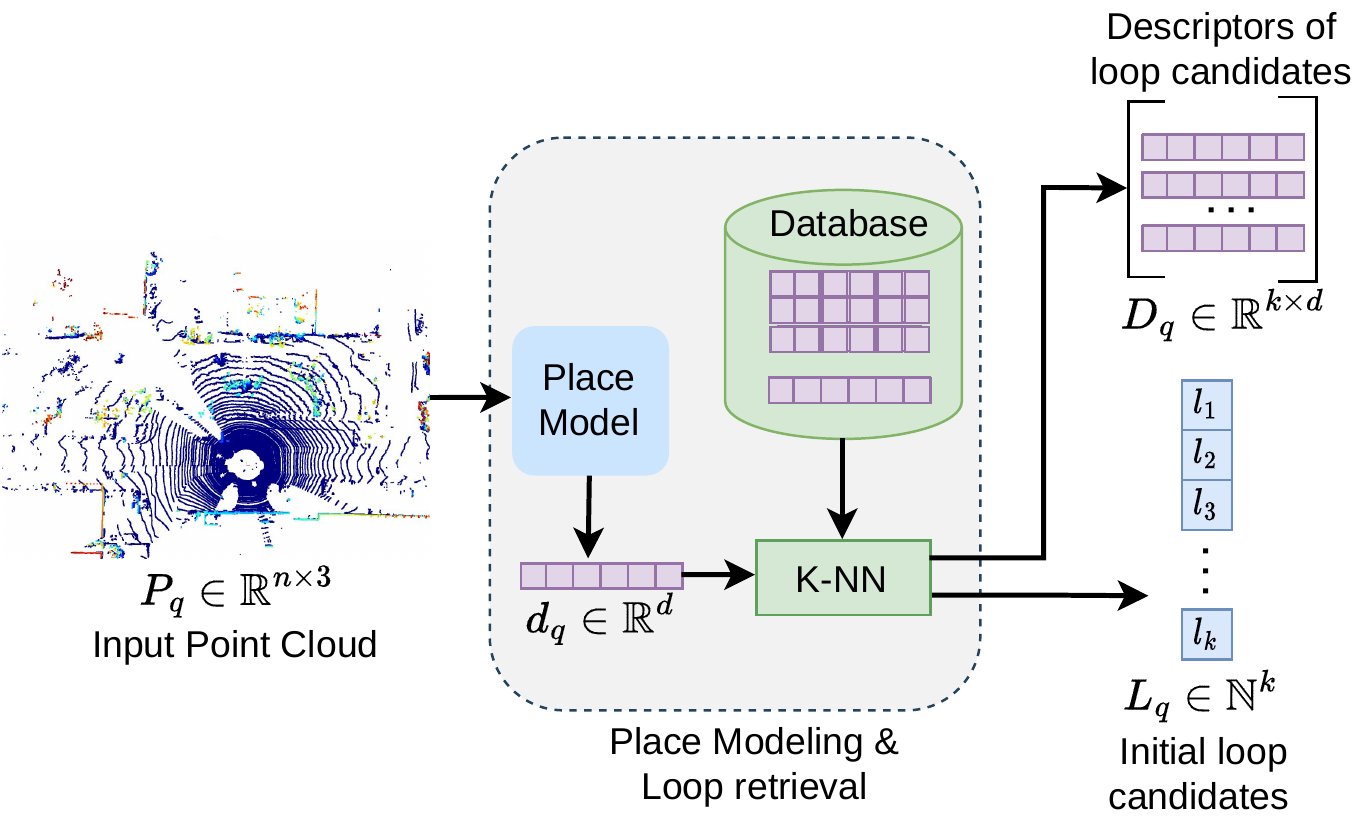}
    \caption{The loop retrieval framework, where the input point cloud $P_q$ is mapped to the descriptor vector $d_q$. The descriptor vector $d_q$ is then used to query the database, after which the $k$ nearest neighbors are returned, serving as loop candidates $L_q$. Both  $L_q$ and the respective descriptors $D_q$ are returned.}
    \label{fig:retrieval}
\end{figure}

In the main framework, shown in Figure~\ref{fig:framework}, the loop retrieval task is performed by the ``Place Modeling \& Loop Retrieval" module, which is detailed in Figure~\ref{fig:retrieval}. This module returns the loop candidates in $L_q$ and their descriptor vectors in $D_q = \{d_i\}_{i=1}^k \in  \mathbb{R}^{k \times d}$, where $k$ is the number of retrieved candidates, each of dimension size $d$.

\subsubsection{Re-Ranking}\label{subsubsec:reranking}

The re-ranking task works on the premise that the initial loop ranking over candidates from $L_q$, contains samples that are not loops (\ie False Positives), in which case we need an approach that is specialized in re-ordering the candidates, ranking higher loop candidates that correspond to real loops and lower loop candidates that correspond to false positives. Mathematically, the re-ranking process can be formulated as follows: given an initial loop ranking over candidates from $L_q$ for a given query point cloud $P_q$, a function $\Theta$ such that:

\begin{equation}
    \Theta:  \mathbb{N}^{k} \rightarrow \mathbb{N}^{k}\mbox{,}
\end{equation}

\noindent maps the initial ranking to a new ranking over the same $k$ candidates, whose corresponding set is $\bar{L}_q  \in \mathbb{N}^{k}$, where the true positives are promoted higher, and the false positives are promoted lower.

\subsubsection{Re-ranking training}
In this work, we resort to the global descriptor-based learning-to-rank approach depicted in Figure~\ref{fig:framework}. In a learning-to-rank setting, the function $\Theta$ is trained. To define the training set, let us define a set of queries $Q\in \mathbb{N}^M$, where $M$ is the number of queries, for each query $q\in Q$ exists a set of loop candidates $L_q \in \mathbb{N}^{k}$, their respective descriptors $D_q \in \mathbb{R}^{k \times d}$ and the ground-truth relevance scores $Y_q = \{y_i=\|p_q - p_{l_i}\|_2 |\forall l_i\in L_q\} \in \mathbb{R}^k$, where $p_q \in \mathbb{R}^3$ and $p_{l_i} \in \mathbb{R}^3$ are the 3D coordinates of the query $q$ and the $l_i$-th loop candidate, respectively. The training set can thus be defined as follows:

\begin{equation}
     \Upsilon = \{(D_q,Y_q) | q \in Q\}\mbox{.}
     \label{eq:traininget}
\end{equation}

Given a training set $\Upsilon$, the goal of training is to find  $\hat{\Theta}$, given by: 
\begin{equation}
    \hat{\Theta} = \argmin(\mathcal{L}(\Theta))\mbox{,}
\end{equation}

\noindent that minimizes the overall loss, given by:
\begin{equation}
    \mathcal{L}(\Theta)= \frac{1}{| \Upsilon |}\sum_{(D_q,Y_q)\in  \Upsilon } l(Y_q, \Theta(D_q))\mbox{,}
\end{equation}

\noindent with $l$ being a ranking loss function of a single query. This loss is computed between the ground-truth relevance scores $Y_q$, and the predicted relevance scores $S_q = \Theta(D_q)$.

\subsection{Transformer-based Re-Ranking}\label{subsec:trer}

This section describes the proposed transformer-based re-ranking approach, TReR, depicted in Figure~\ref{fig:framework}. Instead of following traditional point cloud-based loop re-ranking approaches, which rely predominantly on geometrical cues present in the scene \cite{10065560}, we adopt a more general information re-ranking approach, mainly used in image-related ranking tasks \cite{tan2021instance}. We are furthermore inspired by other works in context ranking, such as \cite{pobrotyn2020context}. However, our approach differs in two ways: (a) the re-ranking model in \cite{pobrotyn2020context} relies on local features, whereas the TReR model relies only on the global descriptors, which allows TReR to be used in applications where it is difficult to extract the local features; (b) the re-ranking model in \cite{pobrotyn2020context} and the TReR model also have different output layers. More specifically, we add a max-polling layer followed by a linear transformation to compute the relative relevance scores used for re-ranking loop candidates.   

\subsubsection{Architecture}\label{subsubsec:architecture}

The proposed transformer-based re-ranker TReR is based on the multi-layer transformer encoders as proposed in \cite{vaswani2017attention}. We let $C$ denote the number of transformer encoders. Pick $i \in \lbrace1,\ldots,C\rbrace$. The $i$-th transformer encoder is outlined in Figure~\ref{fig:framework}. We denote the input of the $i$-th transformer encoder as $Z_{(i)}$. For example, the input of the first encoder (\ie $Z_{(1)}$) is the set of global descriptor vectors $D_q \in \mathbb{R}^{k \times d}$. The multi-layer transformer encoder can thus be defined by the following equations: 

\begin{equation}
    \bar{Z}_{(i+1)} = l_N\left(Z_{(i)} + h_A(Z_{(i)})\right)\mbox{ and}
\end{equation}
\begin{equation}
    Z_{(i+1)} = l_N\left(f(\bar{Z}_{(i+1)})\right)\mbox{,}
\end{equation}

\noindent where $l_N(\cdot)$ is a normalization layer as proposed in \cite{ba2016layer} and $f(\cdot)$ is a multi-layer perceptron network with two linear transformations and a ReLU activation in between, given by: 

\begin{equation}
    f(Z) = \text{max}(0,Z\,W_1+b_1)\,W_2+b_2\mbox{,}
\end{equation}

\noindent where $W_1 \in \mathbb{R}^{d\times d_h}$, $W_2 \in \mathbb{R}^{d_h\times d}$, $b_1 \in \mathbb{R}^{k\times d_h}$, and $b_2 \in \mathbb{R}^{k\times d}$ are weights and biases of linear projections, with $d$ being the input/output dimensionality and $d_h$ the hidden dimensionality. The function $h_A(\cdot)$ is a multi-head attention module with $n$ heads, defined as follows: 

\begin{equation}
    h_A(Z) = W_{AO}\,[h_1, ...,h_j,..., h_n]^{T}\mbox{,}
\end{equation}

\noindent where  $W_{AO} \in \mathbb{R}^{d \times d\,n}$ and $h_j$ is the $j$-th attention head given by the following equation:
\begin{equation}
    h_j = \sigma(\frac{Q\,K^T}{\sqrt{d}}) \, V\mbox{,}
\end{equation}

\noindent with $\sigma(\cdot)$ being the \textit{softmax} function and $d$ is the feature dimensionality, while $Q$, $K$, and $V$ are, respectively, the queries, keys, and values. In this work, the attention mechanism is transformed to self-attention, which results in $Q =Z_{(i)}\,W^Q_j$, $K = Z_{(i)}\,W^K_j$, and $V=Z_{(i)}\,W^V_j$, where $W^Q_j$, $W^K_j$, and $W^V_j \in \mathbb{R}^{d \times d}$ are linear projection matrices of the $j$-th head. 

The $i$-th transformer encoder outputs re-weighted descriptors $Z_{(i+1)} \in \mathbb{R}^{k\times d}$, to which a max-pooling operation is applied alongside the feature dimensions followed by a linear projection:

\begin{equation}
    S_q = \max(Z_{(i+1)})\, W_o + b_o\mbox{,}
\end{equation}

\noindent with $S_q \in \mathbb{R}^k$, whereas $W_o \in \mathbb{R}^{k \times k}$ and $b_o \in \mathbb{R}^k$ encode the weights and biases, respectively. To obtain the final re-ranking, TReR sorts $S_q$ in a descending direction as follows: 
\begin{equation}
    \bar{I}_q = \argsort(S_q)\mbox{,}
\end{equation}

\noindent with $\bar{I}_q\in \mathbb{N}^k$ being a vector of $k$ dimensions, containing the reordering indices, which can be used for transforming the initial ranking of candidates from $L_q$ into the final ranking of candidates from $\bar{L}_q$, as illustrated in Figure~\ref{fig:framework}.

\subsubsection{Training}\label{subsubsec:training}

The proposed re-ranking approach is trained in a supervised regime, where the objective is to maximize the correctness of the relative descriptor order instead of the absolute relevance. As in other ranking works such as~\cite{burges2005learning} and \cite{wang2018lambdaloss}, we adopt the logistic loss, which is a pairwise loss based on the cross-entropy and is given by:

\begin{equation}
    l(Y_q,S_q) = \sum_{y_i > y_j} \text{log}_2(1 + e^{-\sigma(s_i - s_j)})\mbox{,}
\end{equation}

\noindent where $\sigma$ is an adjustable parameter, and $s_i,s_j \in S_q$ and $y_i,y_j \in Y_q$. The advantage of this loss is that it penalizes each out-of-order pair $(i,j)$ that has $y_i>y_j$ but $s_i< s_j$.

\begin{table}[!b]
  \centering
  \caption{The number of frames and loops in the dataset.}
  {\renewcommand{\arraystretch}{1.5}
	\begin{adjustbox}{max width=\columnwidth}
        \begin{tabular}{p{0.30\columnwidth}p{0.10\columnwidth}p{0.10\columnwidth}p{0.10\columnwidth}p{0.10\columnwidth}p{0.10\columnwidth}}
        \noalign{\hrule height 1pt}\toprule	
        Sequence: & 00     & 02     & 05     & 06     & 08 \\
        \midrule
		\midrule
         Length [frames]: &  4051 &  4661 & 2761 &   1101  & 4071 \\
         Queries [loops]:  &  1097 &  450  & 692  &   284   &  521 \\ \bottomrule
        \end{tabular}%
    \end{adjustbox}
    }
  \label{tab:sequence}%
\end{table}%

\begin{table*}[t]
\caption{The loop retrieval results obtained with the baseline models and the respective re-ranking models TReR and $\alpha\text{QE}$ on the dataset. All models were evaluated using a cross-validation approach, tested on a fixed sequence, and trained on the remaining sequences. The presented results are the mean recall scores for the test sequence, $Recall@N$ for $N \in \lbrace 1,5,10\rbrace$, which are denoted as R1, R5, and R10, respectively, computed by using the $N$ most relevant loop candidates. }

{\renewcommand{\arraystretch}{1.2}
\begin{adjustbox}{max width=\textwidth}
\begin{tabular*}{\textwidth}{p{0.2\textwidth}|ccc|ccc|ccc|ccc|ccc}
\hline \noalign{\hrule height 1pt}\toprule	
 & \multicolumn{3}{c|}{00} & \multicolumn{3}{c|}{02} & \multicolumn{3}{c|}{05} & \multicolumn{3}{c|}{06} &  \multicolumn{3}{c}{08}\\ 
 & R1 & R5 & R10 & R1 & R5 & R10 & R1 & R5 & R10 & R1 & R5 & R10 & R1& R5& R10 \\  \midrule \midrule
 LoGG3D-Net & \textbf{0.69}	& 0.75&	\textbf{0.79}&	\textbf{0.55}&	0.58&	0.60&	\textbf{0.68}&	0.76&	0.81&	\textbf{0.91}&	\textbf{0.94}&	0.94&	0.09&	0.25&	0.38\\
LoGG3D-Net + $\alpha\text{QE}$ \cite{radenovic2018fine} & 0.58 &0.71 &	0.78 &	0.48 &	0.58 &	0.60 &	0.59 &	0.72 &	0.79 &	0.75 &	0.91 &	0.95 &	0.08 &	0.25 & 	0.35 \\
 LoGG3D-Net + TReR &  \textbf{0.69} &\textbf{ 0.77} &\textbf{0.79}&	\textbf{0.55}&	\textbf{0.59}&	\textbf{0.61}&	\textbf{0.68}&	\textbf{0.77}&	\textbf{0.82}&	\textbf{0.91}&	\textbf{0.94}&	\textbf{0.95}&	\textbf{0.11}&	\textbf{0.29}&	\textbf{0.42} \\ \hline
ORCHNet & \textbf{0.72} & 0.77 & 0.79 &\textbf{0.53} & \textbf{0.57} & \textbf{0.58} & \textbf{0.65} & 0.71 & 0.73 &\textbf{0.89} & \textbf{0.94} & \textbf{0.95} & \textbf{0.52} & 0.70 & 0.76\\
ORCHNet + $\alpha\text{QE}$ \cite{radenovic2018fine}& 0.60&0.72&0.77&0.40&0.49&0.53&0.52&0.68&0.75&0.74&0.83&0.91&0.52&0.68&0.75 \\ 
ORCHNet + TReR & \textbf{0.72} & \textbf{0.78} & \textbf{0.81} & \textbf{0.53} & \textbf{0.57} & \textbf{0.58} & \textbf{0.65} & \textbf{0.72} & \textbf{0.74} & \textbf{0.89} & \textbf{0.94} & \textbf{0.95} &\textbf{0.52} & \textbf{0.73} & \textbf{0.78}\\\hline
PointNetVLAD & \textbf{0.75} & 0.79 & 0.81 & \textbf{0.51} & 0.60 & 0.65 & \textbf{0.68} & \textbf{0.74} & 0.77 & \textbf{0.92} & \textbf{0.95} & 0.96 & \textbf{0.61} & 0.72 & 0.77\\ 
PointNetVLAD + $\alpha\text{QE}$ \cite{radenovic2018fine} &0.58&0.73&0.78&0.42&0.62&0.66&0.64&0.75&0.79&0.79&0.93&0.94&0.57&0.70&0.78 \\ 
PointNetVLAD + TReR & \textbf{0.75} & \textbf{0.81} & \textbf{0.82} & \textbf{0.51}& \textbf{0.63} &\textbf{0.66} & \textbf{0.68} & \textbf{0.74} & \textbf{0.78} &\textbf{0.92} &\textbf{0.95} & \textbf{0.97} & \textbf{0.61} &\textbf{0.75} & \textbf{0.80}\\ \hline
PoinNetSPoC & \textbf{0.68} & \textbf{0.76} & 0.78 & \textbf{0.47} & 0.56 & 0.59 & \textbf{0.63} & 0.70 & 0.73 & \textbf{0.90} & \textbf{0.93} & 0.94 & \textbf{0.60} & \textbf{0.74} & 0.77 \\
PoinNetSPoC  + $\alpha\text{QE}$  \cite{radenovic2018fine}& 0.29&0.56&0.72&0.28&0.52&0.61&0.36&0.67&0.77&0.53&0.83&0.92&0.40&0.65&0.74\\
PoinNetSPoC + TReR & \textbf{0.68} & \textbf{0.76} &\textbf{0.79} &\textbf{0.47}& \textbf{0.58} & \textbf{0.62} & \textbf{0.63} & \textbf{0.71} & \textbf{0.76} & \textbf{0.90} & \textbf{0.93} & \textbf{0.95} & \textbf{0.60} & \textbf{0.74} &\textbf{0.82}\\ \hline
PointNetGeM & \textbf{0.68} & 0.77 & 0.80 & \textbf{0.49} & 0.60 & 0.64 & \textbf{0.63} & 0.71 & 0.75 &\textbf{0.91} & \textbf{0.95} & \textbf{0.97} & \textbf{0.63} & 0.75 & 0.82 \\
PointNetGeM + $\alpha\text{QE}$  \cite{radenovic2018fine} & 0.23&0.51&0.68&0.29&0.54&0.63&0.32&0.59&0.73&0.59&0.79&0.90&0.55&0.72&0.79\\ 
PointNetGeM + TReR & \textbf{0.68} & \textbf{0.78} & \textbf{0.81} & \textbf{0.49} & \textbf{0.61} & \textbf{0.66} & \textbf{0.63} & \textbf{0.74} & \textbf{0.78} & \textbf{0.91} & \textbf{0.97} & \textbf{0.97} & \textbf{0.63} & \textbf{0.80} & \textbf{0.84}  \\
\hline \noalign{\hrule height 1pt}\toprule	
\end{tabular*}
\end{adjustbox}
}
\label{tab:main}
\end{table*}

\section{EXPERIMENTAL EVALUATION} \label{sec:experiments}

In this section, we describe the experimental and the implementation parts. We also present and discuss the obtained empirical results.

\subsection{Dataset}\label{subsec:dataset}

The proposed approach is evaluated on the KITTI Odometry benchmark \cite{Geiger2012CVPR}, a widely used dataset in various localization and place recognition works. This dataset comprises $22$ sequences with 3D LiDAR, RGB, and RTK-GPS data from urban, highway, and rural environments, representing thus a variety of scenarios that an autonomous vehicle may encounter in real scenarios. Since this work is partly focused on loop retrieval, from within the $22$ original sequences, we only used the sequences that contain revisited segments, which are 00, 02, 05, 06, and 08. Sequence 08 is particularly important because it is the only sequence that has revisits from the opposing direction. As a result, sequence 08 promotes an additional challenge for place recognition methods. In this work, we assume that a loop exists in a given sequence whenever two point clouds were captured from places within a range of 25\,m as proposed in ~\cite{8578568}. Table \ref{tab:sequence} contains the number of loops in each sequence under this assumption and the number of sequence frames.

\begin{figure*}[t]
    \centering
    \begin{subfigure}[b]{0.49\textwidth}
    \includegraphics[width=\textwidth,height=5.50cm,trim={0cm 0cm 0cm 0cm},clip]{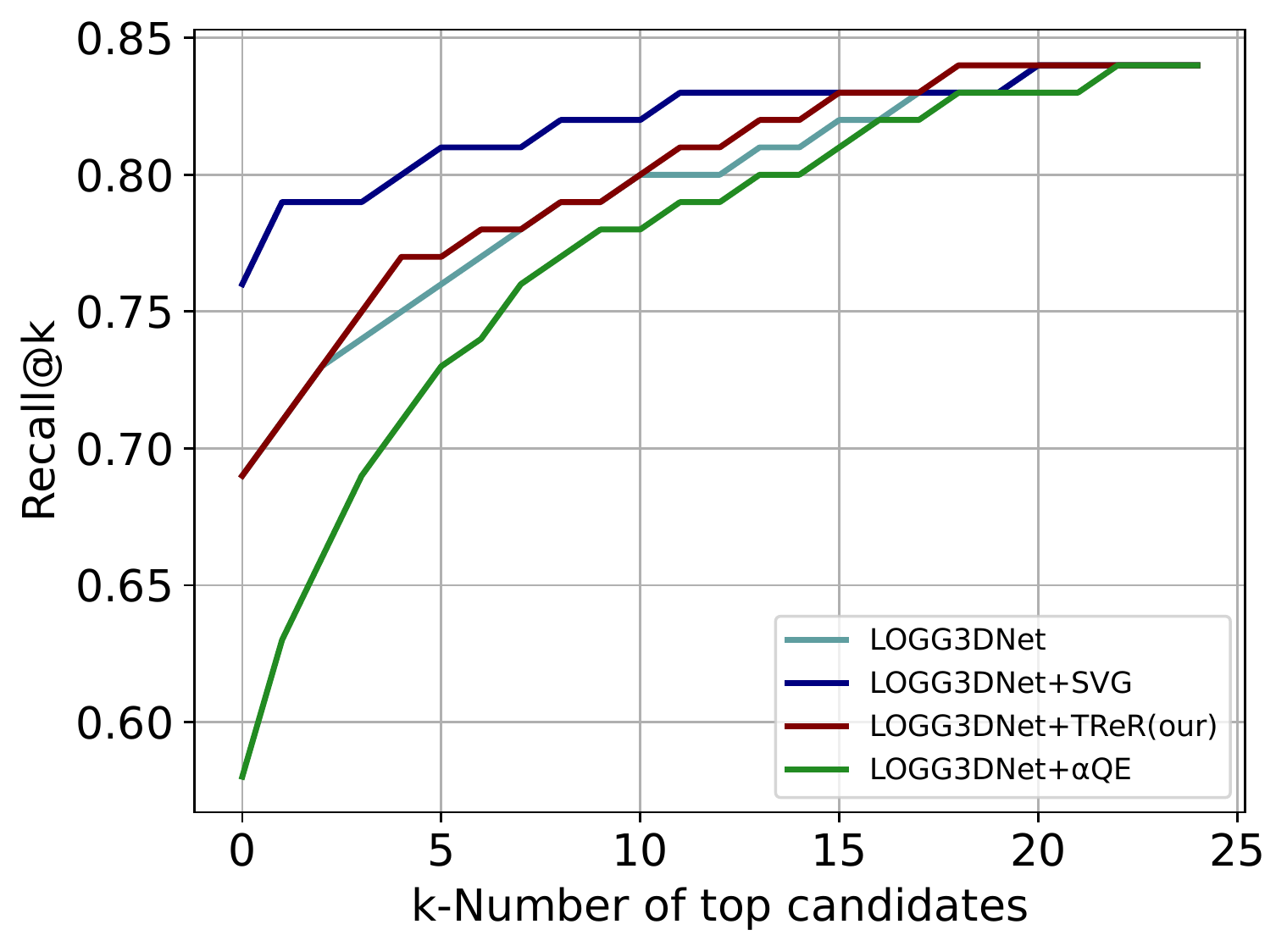}
    \subcaption{}
    \end{subfigure}
    \hfill
    \begin{subfigure}[b]{0.49\textwidth}
    \includegraphics[width=\textwidth,height=5.50cm,trim={0.0cm 0.cm 0cm 0cm},clip]{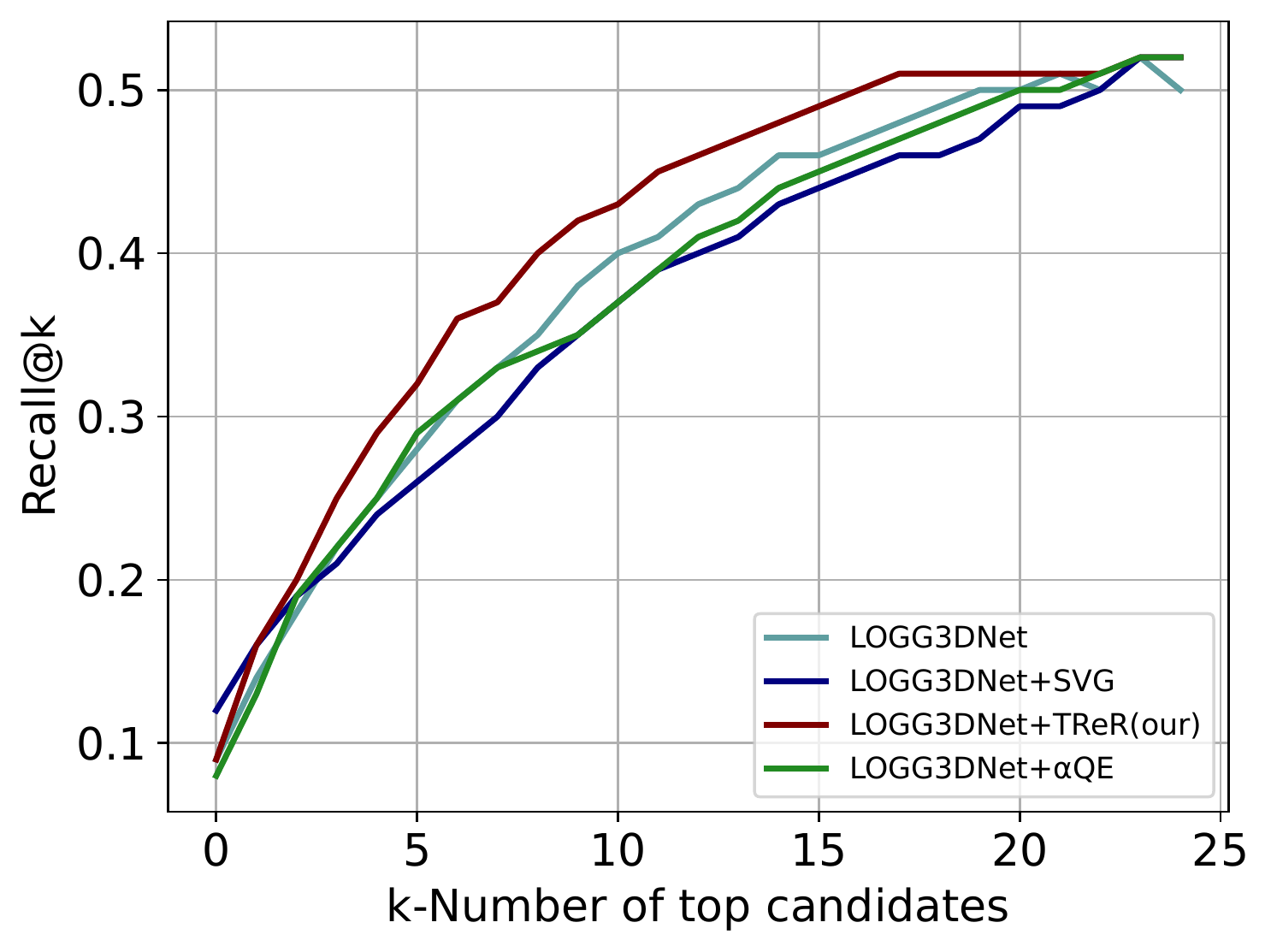}
    \subcaption{}
    \end{subfigure}
    \caption{Comparison between TReR, Spectral Geometric Verification (SGV), and $\alpha$-Query Expansion ($\alpha$QE). LoGG3D-Net was used as a baseline. Key: a) the results for $Recall@k$ on sequence 00; b) the results for $Recall@k$ on sequence 08.}
    \label{fig:recall}
\end{figure*}

\subsection{Experimental Setup}\label{subsec:setup}

All experiments were conducted in Python 3.8 using the PyTorch framework with CUDA 11.6. All models were evaluated using a cross-validation approach: \eg the descriptors of sequence 08 were obtained using a model trained on sequences 00, 02, 05, and 06. As for the training, all models were trained by using the AdamW optimizer with a learning rate (Lr) of 0.0001 and a weight decay (Wd) of 0.0005.

\subsubsection{Baselines}\label{subsubsec:baselines}

TReR was evaluated on global descriptors with 256 dimensions, which were generated by five different models: PointNetVLAD~\cite{8578568}, ORCHNet~\cite{barros2023orchnet}, LoGG3D-Net~\cite{vidanapathirana2022logg3d}, PoinNetSPoC, and PoinNetGeM. PoinNetVLAD and ORCHNet were originally proposed with PointNet as backbone~\cite{qi2017pointnet}. SPoC~\cite{babenko2015aggregating} and GeM~\cite{komorowski2021minkloc3d} were adapted to be implemented on a PointNet-based framework. 
For clouds with less than 10 thousand points, preliminary experiments conducted with PointNetVlAD, ORCHNet,  PoinNetSPoC, and PoinNetGeM indicated that the performance of these models increased with the number of points in the input point clouds. For point clouds with more than 10 thousand points, the performance stagnated. Given these findings, we concluded that a point cloud with 10 thousand points was acceptable for preserving performance and computational demand, which led us to down-sample all point clouds to 10k points using a random sampling approach. The weights of PointNetVlAD~\cite{8578568}, ORCHNet~\cite{barros2023orchnet}, PoinNetSPoC, and PoinNetGeM were trained based on the training protocol described in \cite{uy2018pointnetvlad}. The weights of LoGG3D-Net \cite{vidanapathirana2022logg3d} were pre-trained.

\subsubsection{Re-Ranking}\label{subsubsec:rerankingresults}
We performed experiments on a TReR architecture with one self-attention head (\ie $n = 1$) and one transformer encoder (\ie $C = 1$). As for the input, the descriptors had a size of 256 (\ie $d=256$), the number of maximum candidates was 25 (\ie $k=25$), and $d_h = 512$. With this architecture, TReR had in total 396,033 learnable parameters, which is 4\% of LoGG3D-Net's parameters (8,834,583) and 2\% of PoinNetVLAD's parameters (19,779,145).

\subsubsection{Evaluation}\label{subsubsec:evaluation}

The recall metric is a popular metric for measuring the performance of methods not only in place recognition but also in machine learning. The recall is defined as follows:

\begin{equation}
    Recall = \frac{\mbox{TP}}{\mbox{TP}+\mbox{FN}}\mbox{,}
\end{equation}

\ie the ratio between true positive (TP) and the sum of the true positive and false negative (FN) loops in the ranking. For evaluating the methods, we use $Recall@k$, where $k$ is the number of candidates retrieved from the database.

\subsection{Re-Ranking Evaluation}\label{subsec:rerankevaluation}

We next present and discuss the empirical results. Firstly, we give the results for the baseline methods without re-ranking and with re-ranking, using TReR (our approach) and $\alpha\text{QE}$~\cite{radenovic2018fine}. Secondly, we compare TReR to the spectral geometrical verification (SGV) method~\cite{10065560} and $\alpha\text{QE}$, using as place recognition model LoGGNet3D-Net.

\subsubsection{Global Descriptors Only}\label{subsubsec:global}

We compared the proposed approach TReR with the baselines and $\alpha\text{QE}$ \cite{radenovic2018fine}, which takes top-ranking candidates of a query and generates an updated query. The results are presented in Table~\ref{tab:main}. Overall, they indicate that adding TReR on top of the baseline methods improves the overall performance in terms of $Recall@k$ with $k = [1,5,10]$. TReR tended to have more effect in the middle of the ranking, lifting lower-ranked candidates while leaving unchanged upper-ranked candidates. The results were consistent throughout all models and sequences, indicating the robustness of TReR. On the one hand, TReR had a low impact on ranking estimates that required substantially fewer re-ranking moves, which was the case for sequence 06. On the other hand, TReR had a high impact on ranking estimates that required substantially more re-ranking moves, which was the case for sequence 08. This might be because, as we mentioned, sequence 08 is the most challenging, due to having revisits from the opposing direction. Finally, $\alpha\text{QE}$ underperformed in all experiments. For this reason, we concluded that this approach might not be adequate for this application. 

\subsubsection{Comparison to SGV}\label{subsubsec:comparison}

Furthermore, we compared TReR to the SGV method~\cite{10065560}, which relies on local features and key points to find local feature correspondence. GV-based approaches are usually known for their robustness, whereas approaches that use only global descriptors are usually known for their efficiency. We also observed such performance in our experiments. In Figure~\ref{fig:recall}, we report the results on sequences 00 and 08 for the baseline method LoGGNet3D-Net without re-ranking and LoGGNet3D-Net with re-ranking obtained with TReR, SGV, and $\alpha\text{QE}$. Sequences 00 and 08 represented two interesting edge cases for the methods. On sequence 00, SGV outperformed TReR and $\alpha\text{QE}$, confirming the robustness of GV-based approaches. By comparison, on sequence 08, TReR outperformed SGV and $\alpha\text{QE}$, confirming the efficiency of global-descriptors-only-based approaches. The fact that SGV performed worse than TReR on sequence 08 might be explained by the inability of SGV to find consistencies in pairs of point clouds that are significantly misaligned. In contrast, TReR does not use such local features and key points and, for this reason, performed better than SGV on sequence 08.

\subsection{Runtime Analysis}\label{subsec:runtime}

For each query, TReR re-ranked 25 candidates in a single forward pass, which took approximately 0.3\,s of processing time on average on an AMD Ryzen 9 5900X 12-Core CPU with 64 GB RAM, and 0.001\,s of processing time on average on a GeForce RTX 3090 24GB GPU. These and other runtime results are outlined in Table \ref{tab:latency}. The approach that achieved the lowest performance was $\alpha QE$. Regarding SGV, we were not able to run it on our GPU due to memory constraints. In more detail, SGV required over 27GB of memory when computing key points and local features extracted from the clouds of 10 thousand points, whereas the GeForce RTX3090 had only 24GB for storing such points. Nevertheless, even when accounting just for the CPU time, we could observe that TReR is substantially more efficient than SGV.        

\begin{table}[t]
\caption{The average processing time \CP{in seconds} per query.}
\centering
{\renewcommand{\arraystretch}{1.2}
    \centering
\begin{adjustbox}{max width=\columnwidth}
\begin{tabular}{l|cc}
\hline \noalign{\hrule height 1pt}\toprule	
 Methods& CPU(s) & GPU(s)  \\  \midrule \midrule
TReR & 0.30$\,\pm$\,0.12 & 0.001\,$\pm$0.0005\\ 
SGV &  18\,$\pm$ 2.1 & - \\ 
$\alpha$QE  & 0.00016\,$\pm$0.000002 & 0.00016\,$\pm$0.000002 \\ 
\hline \noalign{\hrule height 1pt}\toprule	
\end{tabular}
\end{adjustbox}
}
\label{tab:latency}
\end{table}

\section{Conclusion}\label{sec:conclusion}

We proposed a 3D LiDAR-based place recognition re-ranking approach called TReR. TReR uses a lightweight transformer-based architecture with global descriptors to re-rank, in a first-retrieve-then-re-ranking paradigm, an initial ranking estimate from a 3D place recognition model. The experimental evaluation was conducted on the KITTI Odometry dataset and highlighted the importance of re-ranking predictions for improving recall over several baseline methods that use no re-ranking and methods such as $\alpha QE$ that use re-ranking. The improvement in recall was consistent over the baselines, showing the robustness of TReR. Similar results were observed when considering the method SGV. TReR improved place recognition model performance, specifically when considering challenging sequences such as sequence 08. Finally, TReR showed a good trade-off between performance and efficiency compared to the SGV method.

\section*{ACKNOWLEDGMENTS}\label{sec:ack}

This work has been partially supported by the project ``GreenBotics" (ref. PTDC/EEI-ROB/2459/2021), funded by Fundação para a Ciência e a Tecnologia (FCT), Portugal. It was also supported by FCT through grant UIDB/00048/2020 and under the Ph.D. grant with reference 2021.06492.BD. Martin Damyanov Aleksandrov was supported by the DFG Individual Research Grant on ``Fairness and Efficiency in Emerging Vehicle Routing Problems" (497791398).

\bibliographystyle{IEEEtran}
\bibliography{root}

\end{document}


\maketitle

\begin{table}[htbp]
\caption{}
{\renewcommand{\arraystretch}{1.2}
\begin{adjustbox}{max width=\textwidth}
\begin{tabular}{l|ccc|ccc|ccc|ccc|ccc|ccc}
\hline \noalign{\hrule height 1pt}\toprule	
 & \multicolumn{3}{|c}{00} & \multicolumn{3}{|c}{02}  & \multicolumn{3}{|c}{05}  &  \multicolumn{3}{|c}{06} &  \multicolumn{3}{|c}{08} &  \multicolumn{3}{|c}{Mean }  \\ 
 & R1 & R5 & R10 & R1 & R5 & R10 & R1 & R5 & R10 & R1 & R5 & R10 & R1 & R5 & R10 & R1 &R5 & R10 \\ \hline
OrchNet & 0.73 & 0.78 & 0,8 & 0.53 & 0.57 & 0.58 & 0.7 & 0.76 & 0.77 & 0.92 & 0.95 & 0.95 & 0.57 & 0.72 & 0.79 & 0.69 & 0.76 & 0.78 \\ 
OrchNet + ReR max & 0.7 & 0.79 & 0.82 & 0.53 & 0.57 & 0.58 & 0.62 & 0.76 & 0.78 & 0.87 & 0.95 & 0.95 & 0.51 & 0.78 & 0.81 & 0.65 & 0.77 & 0.79 \\
OrchNet + ReR CNN & 0.72 & 0.79 & 0.81 & 0.41 & 0.58 & 0.6 & 0.55 & 0.77 & 0.79 & 0.68 & 0.95 & 0.96 & 0.52 & 0.77 & 0.81 & 0.58 & 0.77 & 0.79 \\ 
OrchNet + ReR Wout & 0.26 & 0.79 & 0.81 & 0.42 & 0.57 & 0.6 & 0.64 & 0.77 & 0.81 & 0.87 & 0.94 & 0.95 & 0.5 & 0.77 & 0.82 & 0.54 & 0.77 & 0.80 \\ \hline
PointNetVLAD & 0.75 & 0.79 & 0.82 & 0.52 & 0.61 & 0.65 & 0.73 & 0.77 & 0.8 & 0.94 & 0.94 & 0.94 & 0.63 & 0.73 & 0.79 & 0.71 & 0.77 & 0.80 \\ 
PointNetVLAD + ReR max & 0.68 & 0.81 & 0.83 & 0.38 & 0.64 & 0.68 & 0.29 & 0.8 & 0.83 & 0.74 & 0.94 & 0.94 & 0.54 & 0.78 & 0.83 & 0.53 & 0.79 & 0.82 \\ 
PointNetVLAD + ReR CNN & 0.3 & 0.81 & 0.83 & 0.25 & 0.62 & 0.67 & 0.47 & 0.79 & 0.84 & 0.69 & 0.94 & 0.94 & 0.45 & 0.79 & 0.83 & 0.43 & 0.79 & 0.82 \\
PointNetVLAD + ReR Wout  & 0.73 & 0.8 & 0.82 & 0.32 & 0.63 & 0.68 & 0.47 & 0.8 & 0.83 & 0.82 & 0.94 & 0.95 & 0.48 & 0.77 & 0.83 & 0.56 & 0.79 & 0.82 \\ \hline
PoinNetSpoC & 0.69 & 0.76 & 0.78 & 0.47 & 0.56 & 0.61 & 0.69 & 0.82 & 0.84 & 0.92 & 0.94 & 0.95 & 0.63 & 0.74 & 0.79 & 0.68 & 0.76 & 0.79 \\ 
PoinNetSpoC + ReR max & 0.63 & 0.78 & 0.8 & 0.36 & 0.6 & 0.64 & 0.6 & 0.84 & 0.87 & 0.9 & 0.94 & 0.95 & 0.53 & 0.8 & 0.83 & 0.60 & 0.79 & 0.82 \\ 
PoinNetSpoC + ReR CNN & 0.37 & 0.76 & 0.8 & 0.36 & 0.59 & 0.64 & 0.6 & 0.83 & 0.87 & 0.71 & 0.94 & 0.95 & 0.55 & 0.79 & 0.84 & 0.52 & 0.78 & 0.82 \\ 
PoinNetSpoC + ReR Wout  & 0.23 & 0.77 & 0.8 & 0.36 & 0.58 & 0.63 & 0.5 & 0.83 & 0.86 & 0.71 & 0.94 & 0.95 & 0.49 & 0.8 & 0.84 & 0.46 & 0.78 & 0.82 \\ \hline
PointNetGeM & 0.69 & 0.77 & 0.8 & 0.5 & 0.6 & 0.65 & 0.68 & 0.76 & 0.79 & 0.93 & 0.95 & 0.95 & 0,66 & 0.77 & 0.85 & 0.69 & 0.77 & 0.81 \\ 
PointNetGeM + ReR max & 0.63 & 0.79 & 0.82 & 0.34 & 0.64 & 0.66 & 0.57 & 0.78 & 0.82 & 0.88 & 0.95 & 0.95 & 0.62 & 0.83 & 0.87 & 0.61 & 0.80 & 0.82 \\ 
PointNetGeM + ReR CNN & 0.23 & 0.78 & 0.82 & 0.38 & 0.63 & 0.66 & 0.62 & 0.78 & 0.81 & 0.88 & 0.95 & 0.95 & 0.49 & 0.83 & 0.86 & 0.52 & 0.79 & 0.82 \\ 
PointNetGeM + ReR Wout  & 0.58 & 0.78 & 0.82 & 0.22 & 0.63 & 0.66 & 0.65 & 0.78 & 0.81 & 0.43 & 0.95 & 0.95 & 0.63 & 0.83 & 0.86 & 0.50 & 0.79 & 0.82 \\ \hline
Baseline & 0.72 & 0.78 & 0.80 & 0.51 & 0.59 & 0.62 & 0.70 & 0.78 & 0.80 & 0.93 & 0.95 & 0.95 & 0.62 & 0.74 & 0.81 & 0.69 & 0.76 & 0.80 \\ 
TReR max FC & 0.66 & 0.79 & 0.82 & 0.40 & 0.61 & 0.64 & 0.52 & 0.80 & 0.83 & 0.85 & 0.95 & 0.95 & 0.55 & 0.80 & 0.84 & 0.60 & 0.79 & 0.81 \\ 
TReR CNN & 0.41 & 0.79 & 0.82 & 0.35 & 0.61 & 0.64 & 0.56 & 0.79 & 0.83 & 0.74 & 0.95 & 0.95 & 0.50 & 0.80 & 0.84 & 0.51 & 0.78 & 0.81 \\ 
TReR Wout  & 0.45 & 0.79 & 0.81 & 0.33 & 0.60 & 0.64 & 0.57 & 0.80 & 0.83 & 0.71 & 0.94 & 0.95 & 0.53 & 0.79 & 0.84 & 0.52 & 0.78 & 0.81 \\ \hline
\hline \noalign{\hrule height 1pt}\toprule	
\end{tabular}
\end{adjustbox}
}
\label{}
\end{table}
